\documentclass[preprint,12pt]{elsarticle}

\usepackage{amssymb}
\usepackage{amsmath}

\usepackage{natbib}
\journal{J. Vis. Commun. Image Represent.}
\usepackage[pagebackref,breaklinks,colorlinks]{hyperref}
\usepackage{graphicx}
\usepackage{amsmath}
\usepackage{amssymb}
\usepackage{booktabs}
\usepackage{adjustbox}

\begin{document}

\begin{frontmatter}



\title{3D Surface Reconstruction with Enhanced High-Frequency Details} 

\author[label1] {Shikun Zhang}
\author[label2] {Yiqun Wang\corref{cor1}}
\author[label1] {Cunjian Chen\corref{cor1}}
\author[label2] {Yong Li}
\author[label1] {Qiuhong Ke}
\cortext[cor1]{Corresponding authors.}
\affiliation[label1]{organization={Department of Data Science and AI, Monash University},
           city={Melbourne},
           postcode={3800},
             state={Victoria},
            country={Australia}}

\affiliation[label2]{organization={College of Computer Science, Chongqing University},
            city={Chongqing},
            postcode={401331},
            country={China}}
            

\begin{abstract}
Neural implicit 3D reconstruction can reproduce shapes without 3D supervision, and it learns the 3D scene through volume rendering methods and neural implicit representations. Current neural surface reconstruction methods tend to randomly sample the entire image, making it difficult to learn high-frequency details on the surface, and thus the reconstruction results tend to be too smooth. We designed a method (FreNeuS) based on high-frequency information to solve the problem of insufficient surface detail. Specifically, FreNeuS uses pixel gradient changes to easily acquire high-frequency regions in an image and uses the obtained high-frequency information to guide surface detail reconstruction. High-frequency information is first used to guide the dynamic sampling of rays, applying different sampling strategies according to variations in high-frequency regions. To further enhance the focus on surface details, we have designed a high-frequency weighting method that constrains the representation of high-frequency details during the reconstruction process. Qualitative and quantitative results show that our method can reconstruct fine surface details and obtain better surface reconstruction quality compared to existing methods. In addition, our method is more applicable and can be generalized to any NeuS-based work.
\end{abstract}




\end{frontmatter}



\section{Introduction}
\label{sec1}

Reconstructing a 3D shape from a set of images is an important and challenging task in the field of computer vision \cite{hartley2003multiple}. Recently, the pioneering work NeRF \cite{mildenhall2021nerf} has garnered significant attention since its inception. It combines neural radiance field and volume rendering to achieve high-quality novel view synthesis and implicit representation of 3D shapes. New 3D reconstruction methods based on NeRF were subsequently proposed by NeuS \cite{wang2021neus} and VolSDF \cite{yariv2021volume}, which introduced signed distance function (SDF) to reconstruct smooth surface models. However, all these works have room for improvement in reconstructing surface details.

In general, the quality of surface detail reconstruction can be improved by increasing the number of sampled rays, but this undoubtedly increases the computational consumption. Currently, in order to obtain sufficient sampling information while controlling the computational cost, some methods address these challenges by expanding the sampling area around each ray. For example, as opposed to sampling point-by-point on a ray, Mip-NeRF \cite{barron2021mip} and Tri-MipRF \cite{hu2023tri} sample in the range of the conical table and Gaussian sphere, respectively. This type of methods expands the sampling range, but the sampling is random and does not guarantee that sufficient high-frequency information will be captured. Subsequently, in order to provide more information for the reconstruction of surface details, Guo \cite{guo2022neural} proposes the use of the Manhattan world hypothesis as an additional condition, while Yu \cite{yu2022monosdf} uses pseudo-depth supervision to constrain the reconstruction of texture regions. While this additional information provides additional constraints for surface reconstruction, it is not learned directly for high-frequency details. As an alternative to NeRF, 3DGS \cite{kerbl20233d} has also achieved good success in 3D surface generation tasks. However, 3DGS often faces issues of over-reconstruction during Gaussian densification, where high-variance image regions are covered by only a few large Gaussian blobs. This can hinder the learning of fine local details. Most of the subsequent work based on 3DGS has been devoted to solving the problem of over-reconstruction. Fre-GS \cite{zhang2024fregs} proposes to adjust the training process by constraining the amplitude and phase of high and low frequencies separately in the frequency space. This work also proposed the importance of high-frequency information in the image, but only constrained the relevant parameters in the frequency space and did not further use the high-frequency information to guide the reconstruction process.
\begin{figure*}
    \centering
    \includegraphics[width=1.0\linewidth]{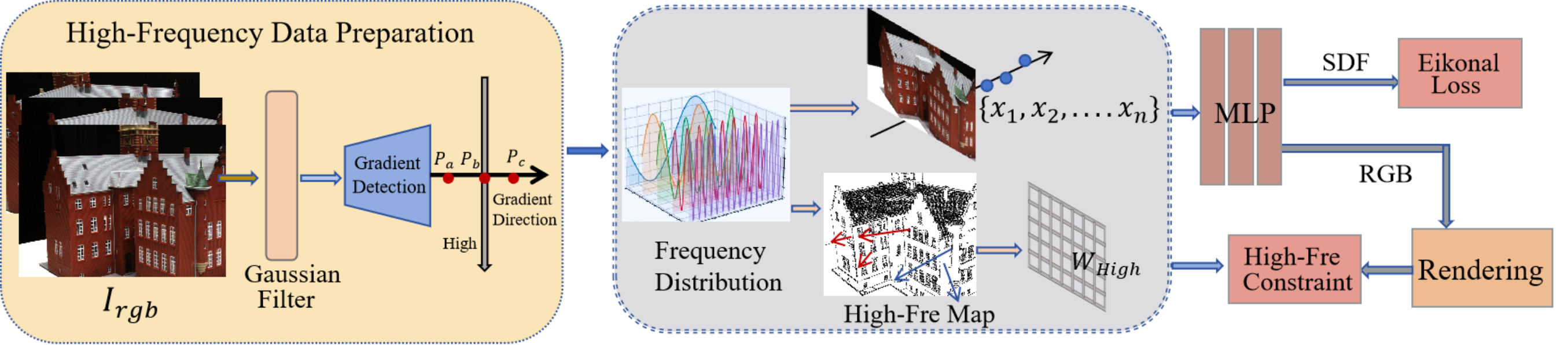}
    \caption{Overview of our proposed Fre-NeuS. We first denoise the input RGB image using Gaussian filtering, then obtain the frequency distribution through the gradient detection module. Based on this distribution, we dynamically adjust the sampling rate to ensure sufficient capture of high-frequency details. We utilize the obtained high-frequency map to create a weight map in pixel space and adjust the optimization of pixels at different frequencies during model training accordingly.}
    \label{fig:network}
\end{figure*}

\begin{figure*}
    \centering
    \includegraphics[width=0.9\linewidth]{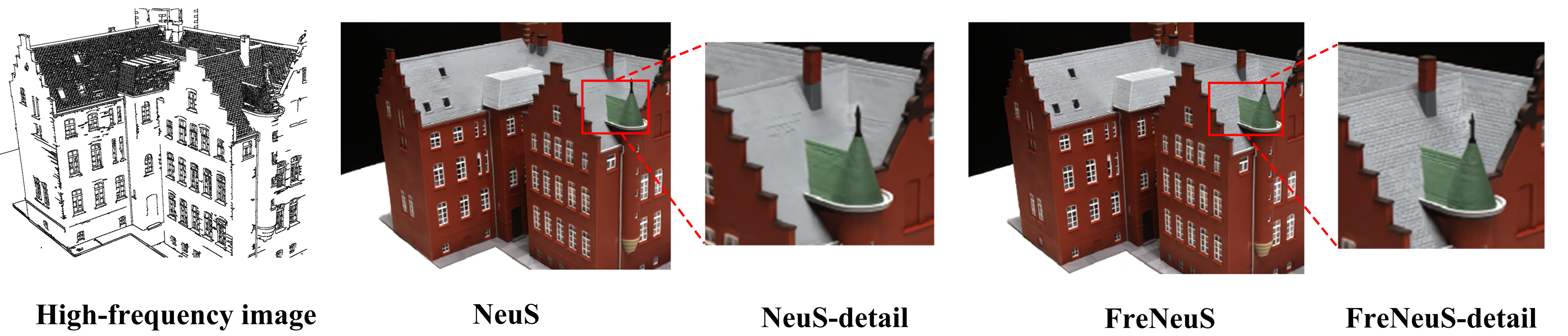}
    \caption{The figure shows the obtained high-frequency image, along with the RGB images rendered by NeuS and Fre-NeuS, respectively. It can be seen that FreNeuS can improve the reconstruction accuracy of texture details by introducing high-frequency information, as demonstrated in features like the roof tile details in the image.}
    \label{fig:edge-compare}
\end{figure*}

Our approach builds on these excellent works and seeks to further improve the quality of the reconstructed surface details. We observe that most of the texture detail information of the target image appears in the high-frequency regions of the surface mutations, which include parts such as depth value changes, color changes and illumination changes. These high-frequency regions can be captured by gradient detection, as shown in Fig. \ref{fig:edge-compare}. The high-frequency image obtained by detection can clearly visualise the details of these surface changes. We further show the view reproduced by the baseline method NeuS \cite{wang2021neus}, as shown in the third column of Fig. \ref{fig:edge-compare}. We can clearly see that the view rendered by the baseline method is unable to reproduce these texture details (e.g., the texture of the roof).

In this paper, we propose a simple but highly effective surface reconstruction method, FreNeuS, which solves the problem of overly smooth reconstructed surfaces by introducing high-frequency information. As shown in Fig. \ref{fig:network}, FreNeuS introduces high-frequency information directly, using pixel gradient variations to detect the target image, thus obtaining an output containing only high-frequency detail variations. In general, low-frequency signals typically encode large-scale features (e.g., global shape and structure, etc.), while high-frequency signals typically encode small-scale features (i.e., local details of a surface). Based on this principle, FreNeuS uses the high-frequency information as a guidance to dynamically adjust the ratio of ray sampling in high and low-frequency regions. Compared to random sampling, Fre-NeuS dynamically increases the number of samples in detailed regions, optimizing the overall sampling distribution. In addition, we propose a constraint mechanism for high-frequency information. By constraining the high-frequency signals in pixel space, we seek to minimize the difference between the details of the rendered image and the corresponding ground truth. The experimental results demonstrate that FreNeuS enhances the representation of high-frequency texture details (e.g., roof textures, etc.), as shown in Fig.~\ref{fig:edge-compare}. Further, the overall quality of surface reconstruction is also improved, as shown in Table \ref{table:dtu}. 

The main contributions of this work are summarized here:
\begin{itemize}
    \item We propose a high-frequency enhancement module that reconstructs surface detail textures by introducing high-frequency information as additional input. 
\end{itemize}
\begin{itemize}
    \item We propose an innovative dynamic sampling strategy based on high-frequency information, which ensures sufficient sampling of local details.
\end{itemize}
\begin{itemize}
    \item A high-frequency information constraint mechanism is designed to enhance the learning of high-frequency signals in pixel space, effectively minimizing surface detail errors.
\end{itemize}
\begin{itemize}
    \item Experiments on benchmark datasets demonstrate that Fre-NeuS excels in surface reconstruction and significantly enhances local detail accuracy. Furthermore, the method integrates seamlessly into any NeuS-based framework, improving texture detail reconstruction.
\end{itemize}

\section{Related Work}
\label{sec2}
Multi-view 3D reconstruction has been a challenging task in the field of computer vision. Traditional multi-view 3D reconstruction methods usually use explicit geometry to express 3D structures, which can be classified into voxel-based methods \cite{broadhurst2001probabilistic,de1999poxels,izadi2011kinectfusion,kutulakos2000theory,niessner2013real,seitz1999photorealistic} and point-based methods \cite{barnes2009patchmatch,galliani2016gipuma,schonberger2016structure,schonberger2016pixelwise} according to the representation. The voxel-based approach has high memory overhead, while the point-based approach requires additional consideration of missing point clouds. 
Recently, neural implicit surface-based reconstruction methods have been proposed, which provide new ideas for novel view synthesis and surface reconstruction tasks. Subsequently, the 3D Gaussian Splatting technique for 3D scene modelling has also gained widespread support from researchers due to its efficient training speed and superior rendering quality. Here, we provide a brief review of these two methods.


\textbf{Neural surface reconstruction.} Neural surface reconstruction is based on the neural implicit representation of surfaces, which uses successive neural implicit functions to reconstruct shapes from multi-view images. Neural Radiance Fields (NeRF) is a specific neural field application used mainly for 3D scene reconstruction and image synthesis. In the work of NeRF \cite{mildenhall2021nerf}, it uses volume rendering to map a 3D density field and a 3D oriented colour field to a 2D image. Subsequent work, such as VolSDF \cite{yariv2021volume} and NeuS \cite{wang2021neus}, also incorporates occupancy functions or signed distance functions into the volume rendering equation. Since the implicit function can be regularized by the Eikonal loss, the reconstructed surface can maintain smoothness. They also use MLP to model 3D scenes of multi-view 2D images and can generate new views with excellent multi-view consistency. Subsequently, many variants of NeuS have been designed in order to improve the accuracy of surface detail reconstruction. To further improve surface fidelity, HF-NeuS \cite{wang2022hf} introduces an additional MLP for modelling displacement fields to learn high-frequency details. In contrast to these methods, we believe that the reconstruction of local details can be more directly constrained by using high-frequency signals as guiding information, and therefore we propose the new method FreNeuS. 

\textbf{Gaussian Splatting}. Once proposed, the Gaussian Splatting technique has attracted much attention for its real-time rendering speed and excellent new view generation. The 3D Gaussian Splatting \cite{kerbl20233d} introduces an anisotropic 3D Gaussian distribution and an efficient microscopic splash map, and then it utilises SfM \cite{snavely2006photo} to estimate a dominant sparse point cloud from a set of multiple images. Therefore, compared with other 3D scene modelling methods, Gaussian Splatting has a great advantage in terms of training speed. However, during Gaussian densification, too many 3D Gaussians tend to cause blurring of the reconstructed surface, making it difficult to reproduce rich surface details. Based on this, several works \cite{jiang2023gaussianshader,gao2023relightable} extended 3DGS by modelling the normals as an additional attribute of the 3D Gaussian primitive. This year the proposal of 2D Gaussian \cite{huang20242d} attracted scholars' attention again. The method defines normals by using 2D Gaussian primitives to represent the tangent space of 3D surfaces, aligning them more closely with the underlying geometry. This method reconstructs a smoother surface compared to 3DGS. Both the method based on neural implicit surfaces and the method based on Gaussian Splatting techniques are outstanding representatives of current surface reconstruction tasks. Therefore, we compare FreNeuS with the excellent work mentioned above.

\begin{figure*}
    \centering
    \includegraphics[width=1.0\linewidth]{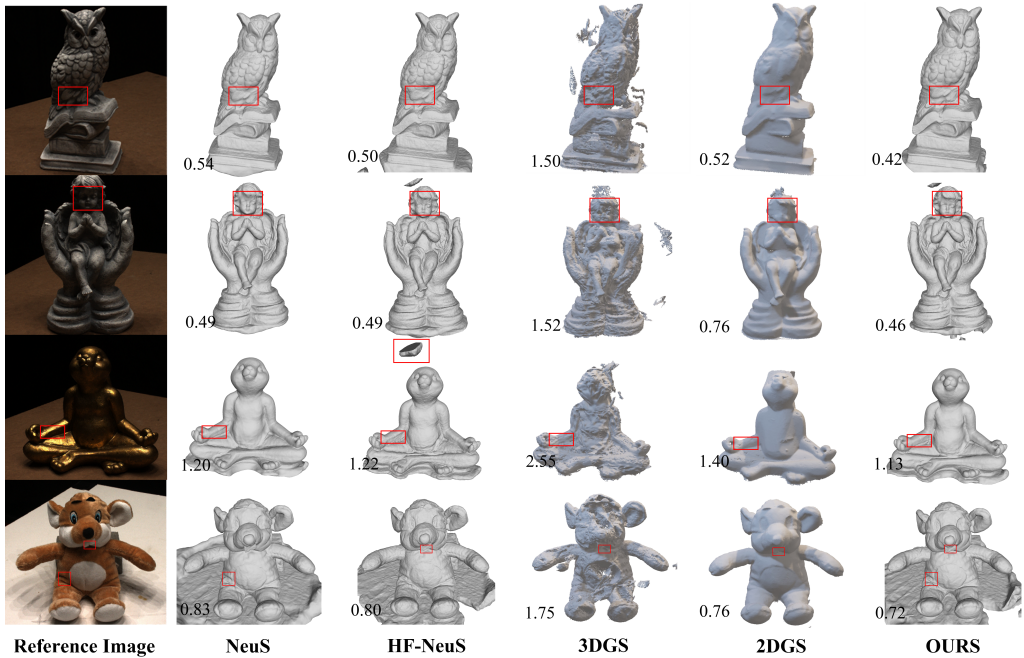}
    \caption{Visualisation comparisons on different scenes of the DTU dataset. First column: reference images. Second to fifth columns: NeuS, HF-NeuS, 3DGS, 2DGS, and OURS. Chamfer distance errors are labelled in the lower left corner, with smaller metrics being better. Our method achieves minimal error while effectively capturing surface details. For instance, in the fourth row of the bear scene, Fre-NeuS successfully reproduces both the tooth gap and the distinct leg folds.}
    \label{fig:dtu-compare}
\end{figure*}

\section{Method}
\label{sec:method}
Given a set of images and their corresponding camera positions, the goal of FreNeuS is to reconstruct a 3D scene represented by signed distance functions (SDF). FreNeuS aims to enhance surface detail reconstruction by leveraging high-frequency information as guidance. In this section, we first describe how to detect the target image by analyzing pixel gradient changes, which allows us to obtain high-frequency information. Next, we demonstrate how the sampling ratio can be dynamically adapted based on the extracted high-frequency information. Finally, we present the design of the high-frequency information constraint mechanism.

\subsection{High-frequency Detection}
Compared to the original image, high-frequency maps offer a clearer representation of surface detail changes. We propose a high-frequency enhancement module to capture frequency information. It detects high and low frequencies by identifying changes in gradient intensity. Points in high-frequency regions are considered high-frequency points, including areas of depth discontinuities, scene lighting variations, and other critical transitions. These points reflect surface inconsistencies and highlight regions that require attention during the reconstruction process.

In the process of extracting high-frequency maps, the first Gaussian filter $G(x,y)$ is used to remove the noise from the input image. Next, we compute the gradient and orientation. $G_{x}(x,y)$ and $G_{y}(x,y)$ denote the gradient values in horizontal and vertical directions respectively. Accordingly, the gradient intensity and direction can be calculated as:
\begin{align}
\label{eq1}
M(x, y)=\sqrt{G_{x}^{2}(x, y)+G_{y}^{2}(x, y)}\ ,\notag \\
\theta(x, y)=\tan ^{-1}\left(\frac{G_{y}(x, y)}{G_{x}(x, y)}\right).
\end{align}

Here, $M(x,y)$ denotes the gradient intensity of the pixel point and $\theta(x,y)$  denotes the gradient direction of the pixel point. Then, boundary tracing is performed based on the grayscale threshold to identify the high-frequency regions of the image. After obtaining the high-frequency distribution of the image, it is fed into the network to guide the subsequent surface reconstruction task.

\subsection{High-frequency Dynamic Sampling}
Most NeuS-based approaches utilize random sampling as their strategy, where a batch of rays is randomly selected from the set of candidate pixels in the target image. Random sampling is simple to operate and ensures the diversity of batches in each sampling. However, in the reconstructed scene image, smooth regions without texture details account for a large proportion, and if a random sampling strategy is used to collect rays from the target image, it naturally leads to under-sampling of regions with high-frequency detail variations. This is not conducive to recovering fine details on the object surface. We observe that most of these details are concentrated in high-frequency regions. To address this, we propose a novel high-frequency-guided ray sampling strategy that dynamically adjusts sampling ratios across different regions based on high-frequency information.

First, we determine the high-frequency pixel set based on the frequency distribution of the target image, as shown in Eqn.~\eqref{eq2}:
\begin{equation}
\label{eq2}
I_{H}, I_{L}=\left\{\begin{array}{l}
P_{e}>0, P \in I_{H} \\
P_{e}<=0, P \in I_{L}.
\end{array}\right.
\end{equation}

In Fig. \ref{fig:network}, the red arrows on the high-frequency map represent high-frequency pixels, while the blue arrows indicate low-frequency pixels. Here, $P_{e}$ denotes the pixel value of the point pixel point $P$ in the high-frequency map, and the high-frequency pixel set $I_{H}$ of the target image is obtained after the above processing. Correspondingly, the set of low-frequency region pixels in the target image is $I_{L}$.

Then, instead of setting fixed sampling weights, we randomly sample each target image according to the proportion of high-frequency and low-frequency regions in it. This approach allows for dynamic adjustment of sampling for each target image, ensuring that sampling is both balanced and targeted. Mathematically, this could be described as:
\begin{equation}
\label{eq3}
\begin{array}{c}
H_{\text {num}}=w * B, \\
L_{\text {num}}=(1-w) * B.
\end{array}
\end{equation}

Specifically, the ratio of high-frequency pixels to low-frequency pixels is used as the sampling weight $w$, i.e., \(w = I_{H}/I_{L}\), and $B$ represents the total number of rays sampled at each step during the training process. The $H_{num}$ in Eqn.~\eqref{eq3} denotes the number of samples in the high-frequency pixel set, and $L_{num}$ denotes the number of samples in the low-frequency pixel set. Then according to the number of samples, random samples are taken in $I_{H}$ and $I_{L}$, respectively. The advantage of this approach is that it preserves the randomness of sampling while ensuring a higher sampling rate in detail areas through constraints based on high-frequency information. This is beneficial for reproducing detailed textures in the reconstruction.

\subsection{High-frequency Constraint Mechanisms}
As mentioned earlier, it is crucial to distinguish the contribution of high-frequency regions from that of low-frequency regions. Previously, we employed dynamic sampling to ensure adequate representation of high-frequency regions, thereby improving the learning of high-frequency detail textures. Furthermore, by looking at the rendered view of the baseline NeuS, we found that the detail regions in the view were largely blurred and still lacked finer texture. As shown in Fig. \ref{fig:edge-compare}, the baseline NeuS rendered view is noticeably lacking in details such as roofs, windows, etc. compared to the original image. This can be summarized as: the rendered view appears overly smooth in areas with sharp texture details.

Based on this, we believe that in addition to enhancing the sampling of high-frequency regions, we can add limiting constraints to these regions based on high-frequency map. Specifically, we add additional constraints to the high-frequency signals in pixel space during the rendering process, as shown in Eqn. \eqref{eq4}:
\begin{equation}
\label{eq4}
\begin{array}{c}
W_{\text {High }}=\left(w_{\text {High }}^{1}, w_{\text {High }}^{2}, \ldots . ., w_{\text {High }}^{n}\right), \\
w_{\text {High }}^{i}=\left\{\begin{array}{ll}
a, & P \in I_{H} \\
b, & P \notin I_{H}.
\end{array}\right.
\end{array}
\end{equation}

First, we assess whether the sampled ray falls within a high-frequency region based on pixel values, and then assign weights to the ray according to this assessment. The weight of the ray in the high-frequency region is $a$, and the weight of the ray in the low-frequency region is $b$. The $W_{\text {High }}$ denotes the set of weights of all sampled rays, and $w_{\text {High }}^{i}$ denotes the weight of the $ith$ pixel.

After obtaining the set of pixel weights, these weights will be used to constrain the expression of high-frequency pixels during the training process, as illustrated in Eqn. \eqref{eq5}:
\begin{equation}
\label{eq5}
L_{\text {Frecolor }}=\frac{1}{|S|} \sum_{s \in S}\left\|\widehat{C}_{s}-C_{s}\right\| W_{\text {High }},
\end{equation}

where $\widehat{C}_{s}$ is the ground truth color, the volume rendering color is $C_{s}$, and $S$ is the total number of sampled pixel sets. After constraining the high-frequency pixels, the model further improves the reconstruction quality of the surface texture details. Specific results are shown in Table \ref{table:ablation} and Fig. \ref{fig:Ablation}.

\subsection{Optimization}
In addition to using high-frequency constraints to limit the rendered color, we also use Eikonal loss \cite{gropp2020implicit}. Eikonal loss is a regularised loss on the set of sampled points $I$ that constrains the implicit function, as shown in Eqn. \eqref{eq6}:
\begin{equation}
\label{eq6}
L_{\text {reg }}=\frac{1}{|I|} \sum_{i \in I}\left[\left(\| \nabla S\left(x_{i}, y_{i}, z_{i} \|\right)\right]-1\right..
\end{equation}
We train our network using both the rendering loss $L_{\text {egcolor }}$ and Eikonal loss $L_{\text {reg }}$ functions as shown below: 
\begin{equation}
\label{eq7}
L_{\text{total}} = L_{\text{reg}} + \lambda L_{\text{Frecolor}}.
\end{equation}

\section{Experiments}
\label{sec:Experiments}
\textbf{Dataset}. We first conducted experiments on the $DTU$ dataset \cite{jensen2014large}, which shows all real scenes. $DTU$ is a multi-view stereo dataset, where each scene consists of $49$ or $64$ views with a resolution of $1600 \times 1200$. We followed the previous method to select the same $15$ scene data from it. We further selected six challenging scenes from other datasets, the NeRF synthetic dataset \cite{mildenhall2021nerf}. The NeRF synthetic dataset \cite{mildenhall2021nerf} has an image resolution of $800 \times 800$ and provides $100$ views per scene. We chose this dataset to analyse the reconstruction of high-frequency details.

\textbf{Baseline}. We compare FreNeuS with the following five state-of-the-art baselines: (1) NeuS \cite{wang2021neus} is the most relevant baseline for our work, and much of the current work is based on NeuS. (2) VolSDF \cite{yariv2020multiview} is the work related to NeuS, which is also a very popular framework, and it performs well. (3) HF-NeuS \cite{wang2022hf} is one work based on NeuS, which proposes to construct SDF functions using displacement functions and achieves good results. (4) In addition to the NeuS-based work, we also selected the current state-of-the-art Gaussian Splatting-based work, 3DGS \cite{Huang2DGS2024} and 2DGS \cite{Huang2DGS2024}. For all methods, we used a threshold of $25$ to extract surfaces for comparison. 

\textbf{Evaluation metrics}. To evaluate the quality of the reconstruction, we followed previous work and used the evaluation metrics Chamfer distance (lower values are better). Following the official protocol, the background in the DTU dataset is not part of the ground truth surface, so we followed the previous method to remove the background when calculating the chamfer distance. NeRF synthetic dataset \cite{mildenhall2021nerf} does not have a background, so we calculated the chamfer distance between the ground truth shape and the reconstructed surface.

\begin{table*}[ht]
\centering
\begin{adjustbox}{max width=\textwidth}
\small 
\begin{tabular}{lcccccccccccccccc}
\toprule
\text{Method} & 24 & 37 & 40 & 55 & 63 & 65 & 69 & 83 & 97 & 105 & 106 & 110 & 114 & 118 & 122 & \text{Mean} \\
\midrule
\text{NeRF} & 1.90 & 1.60 & 1.85 & 0.58 & 2.28 & 1.27 & 1.47 & 1.67 & 2.05 & 1.07 & 0.88 & 2.53 & 1.06 & 1.15 & 0.96 & 1.49 \\
\text{VOLSDF} & 1.14 & 1.26 & 0.81 & 0.49 & 1.25 & 0.70 & 0.72 & 1.29 & 1.18 & \textbf{0.70} & 0.66 & \textbf{1.08} & 0.42 & 0.61 & 0.55 & 0.86 \\
\text{NeuS} & 1.00 & 1.37 & 0.93 & 0.43 & 1.10 & 0.65 & \textbf{0.57} & 1.48 & \textbf{1.09} & 0.83 & 0.52 & 1.20 & 0.35 & 0.49 & 0.54 & 0.84 \\
\text{HF-NeuS} & 0.76 & 1.32 & 0.70 & 0.39 & 1.06 & 0.63 & 0.63 & \textbf{1.15} & 1.12 & 0.80 & \textbf{0.52} & 1.22 & 0.33 & 0.49 & 0.50 & 0.77 \\
\text{3DGS} & 2.14 & 1.53 & 2.08 & 1.68 & 3.49 & 2.21 & 1.43 & 2.07 & 2.22 & 1.75 & 1.79 & 2.55 & 1.53 & 1.52 & 1.50 & 1.96 \\
\text{2DGS} & \textbf{0.48} & 0.91 & \textbf{0.39} & 0.39 & 1.01 & 0.83 & 0.81 & 1.36 & 1.27 & 0.76 & 0.70 & 1.40 & 0.40 & 0.76 & 0.52 & 0.80 \\
\text{FreNeuS (ours)} & 0.81 & \textbf{0.82} & 0.74 & \textbf{0.37} & \textbf{0.97} & \textbf{0.56} & 0.58 & 1.40 & 1.19 & 0.72 & 0.53 & 1.13 & \textbf{0.32} & \textbf{0.46} & \textbf{0.42} & \textbf{0.73} \\
\bottomrule
\end{tabular}
\end{adjustbox}
\caption{Quantitative results on DTU (the header’s numbers denote scene IDs). It can be seen that our method minimises the average error across all scenarios.}
\label{table:dtu}
\end{table*}

\textbf{Implementation details}. Our overall network framework follows the structure of NeuS and uses an $8$-layer MLP structure. We trained the network using Adam with a learning rate of $5e-4$ and we used an NVIDIA RTX A6000 48GB GPU. For training, a total of $512$ rays are sampled in each batch, the sampling region is assigned according to section \ref{sec:method}, and then $64$ points are uniformly sampled on the rays. Then we calculate the SDF value and its gradient, and fine sampling is performed again based on the SDF value. The sampling of the view is completed by following the strategy from coarse to fine. In addition, we set $\lambda$ to $1.2$.

\subsection{Comparison}
In Table \ref{table:dtu}, we compare the chamfer distance error of FreNeuS with 3DGS, 2DGS, and several other NeuS-based methods on the dataset DTU. We can observe that FreNeuS achieves the smallest error in most of the scenes, and it also achieves the best average performance in all scenes. Furthermore, we show the visualisation of some of the scenes in Fig. \ref{fig:dtu-compare}. Compared to other methods, the FreNeuS reconstruction of the surface is more noise-resistant and more accurate in reconstructing the local details of the surface. For example, for the bronze statue in the third row, the FreNeuS reconstructed surface has clearer facial lines and less noise on the stomach. In the fourth row of the bears, Fre-NeuS accurately reproduces the details of the teeth gaps, and the leg folds are more clearly defined. In contrast, NeuS and HF-NeuS fail to capture these finer details as effectively.

\begin{figure*}
    \centering
    \includegraphics[width=0.95\linewidth]{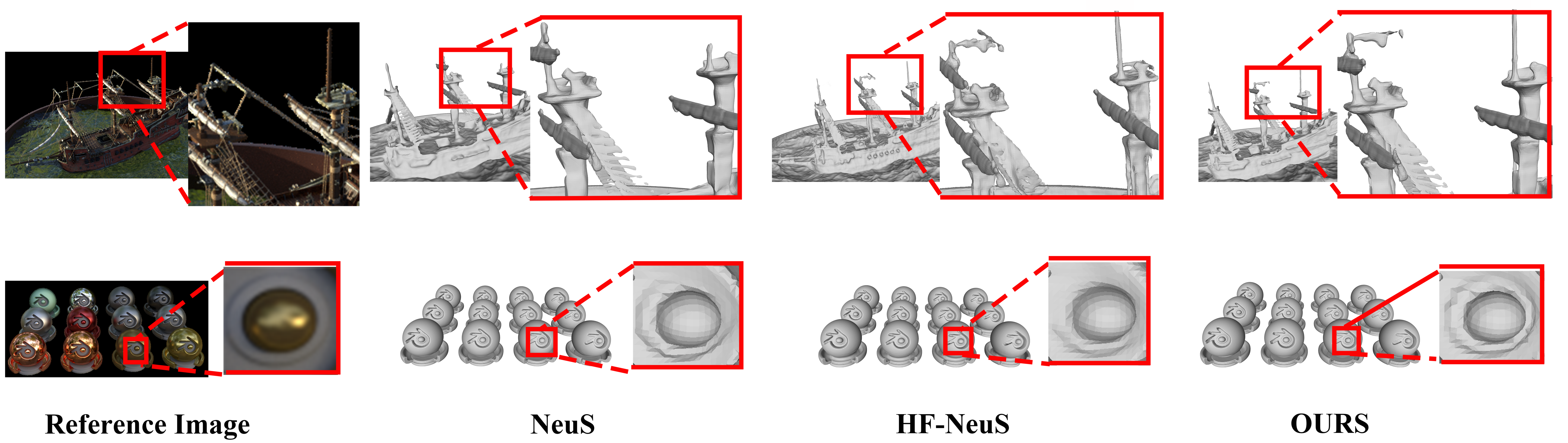}
    \caption{Qualitative evaluation on the Ship and Material scenes. First column: reference images. Second to the fourth column: NeuS, HF-NeuS, and OURS. Compared to other methods, the inner circle of the sphere reconstructed by FreNeuS is more visible and the sail texture is much clearer.}
    \label{fig:nerfsy}
\end{figure*}

\begin{table}[ht]
\centering
\begin{adjustbox}{scale=0.8}
\small 
\begin{tabular}{l@{\hskip 2pt}ccccccc}
\toprule
\text{Method} & \text{Chair} & \text{Ficus} & \text{Lego} & \text{Materials} & \text{Mic} & \text{Ship} & \text{Mean} \\
\midrule
\text{VOLSD} & 1.26 & 1.54 & 2.83 & 1.35 & 3.62 & 2.92 & 2.37 \\
\text{NeuS} & 0.74 & 1.21 & 2.35 & 1.30 & 3.89 & 2.33 & 1.97 \\
\text{HF-NeuS} & 0.69 & 1.12 & \textbf{0.94} & 1.08 & \textbf{0.72} & 2.18 & 1.12 \\
\text{FreNeuS (ours)} & \textbf{0.45} & \textbf{0.63} & 1.30 & \textbf{0.13} & 2.7 & \textbf{1.50} & \textbf{1.11} \\
\bottomrule
\end{tabular}
\end{adjustbox}
\caption{Quantitative results on the NeRF-synthetic dataset.}
\label{table:nerfsy}
\end{table}
Most of the scenes in the DTU dataset are smooth surfaces where high-frequency details are not apparent. Therefore, we selected six challenging datasets from the NeRF synthetic dataset \cite{mildenhall2021nerf} for comparison. As shown in Table \ref{table:nerfsy}, we show the results of FreNeuS and other methods in the six synthetic data. Based on the evaluation results, it can be seen that FreNeuS achieves the smallest error in the average results of all the scenes. Among them, the performance in the Mic scene has a greater improvement compared to the baseline NeuS, but not as good as HF-NeuS. we analysed that it is because of the lack of high-frequency lines in the stent portion of the Mic itself, which has less impact on the global after adding the high-frequency constraints. In addition, in order to visualise the reconstruction effect on local details, we show a visual comparison of FreNeuS with the other two methods in Fig. \ref{fig:nerfsy}. In the first row of the ship scene, Fre-NeuS reconstructs the cut-out details of the sails and the long mast section. In comparison, NeuS and HF-NeuS fail to capture the long mast and the sail cut-outs, respectively. In the second row of material scenes, the segmentation of the ball and base by HF-NeuS is not clear, and NeuS does not clearly reconstruct the small inner-most circle of the ball.

\subsection{Ablation Study}
\subsubsection{Design verification}
\begin{figure*}
    \centering
    \includegraphics[width=0.9\linewidth]{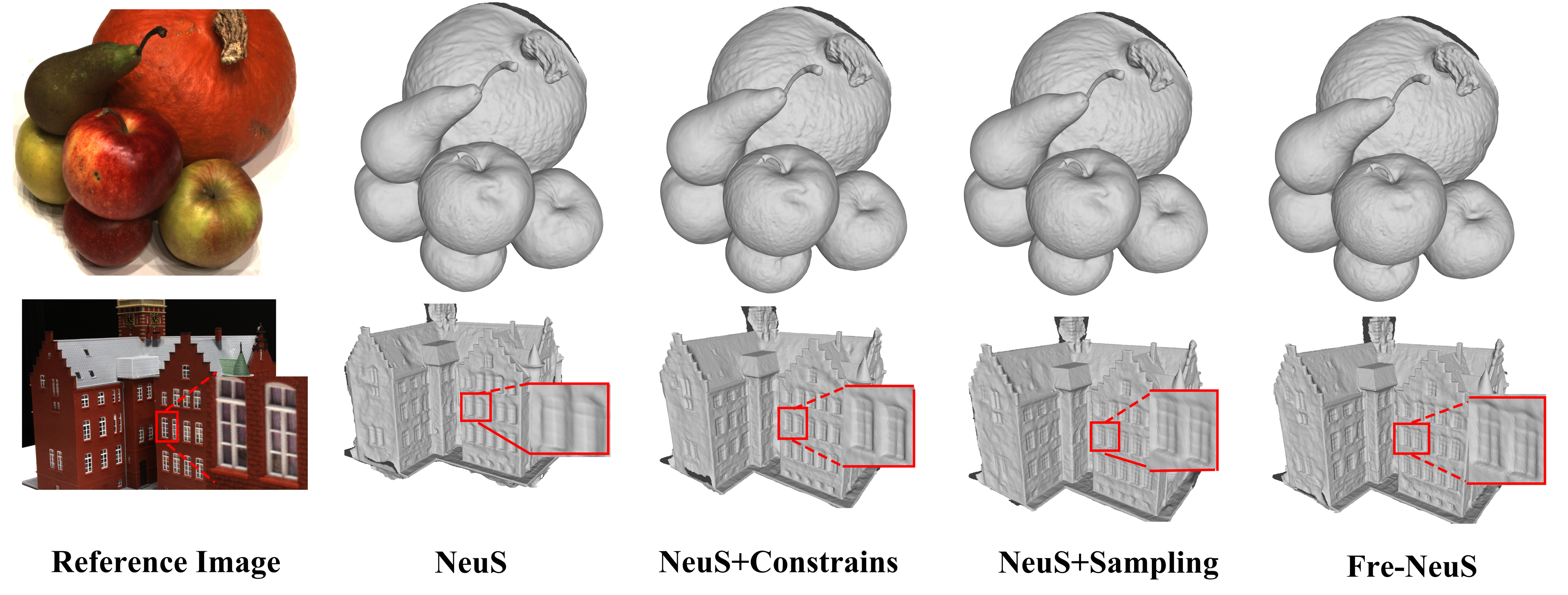}
    \caption{Visualisation of ablation experiments to verify the performance of each module. First column: the reference image. Second to fifth columns: NeuS, use of high-frequency constraints, use of high-frequency dynamic sampling strategies, and FreNeuS. It can be observed that as the two modules are added, the reconstructed surface details become clearer and more refined.}
    \label{fig:Ablation}
\end{figure*}

We conducted quantitative experiments on the DTU dataset to assess the impact of various modules on reconstruction results, including the high-frequency dynamic sampling and high-frequency constraint mechanisms. In Table \ref{table:ablation}, ‘Base’ denotes the baseline method, i.e., NeuS; ‘+sampling’ denotes the high-frequency guided ray sampling strategy; ‘+Fre-constraints ‘ denotes the high-frequency constraints mechanism. We use ‘FreNeuS’ as the complete structure of our proposed method. Based on the data in Table \ref{table:ablation}, it can be seen that both of our proposed modules improve the fidelity of the reconstruction. In addition, we show the view comparison in Fig. \ref{fig:Ablation}. This also demonstrates that FreNeuS is effective in reconstructing surface details using high-frequency information.

\begin{table*}[ht]
\centering
\begin{adjustbox}{max width=\textwidth}
\small 
\begin{tabular}{lcccccccccccccccc}
\toprule
\text { Method } & 24 & 37 & 40 & 55 & 63 & 65 & 69 & 83 & 97 & 105 & 106 & 110 & 114 & 118 & 122 & \text { Mean } \\
\midrule
\text { Base } & 1.00 & 1.37 & 0.93 & 0.43 & 1.10 & 0.65 & \textbf{0.57} & 1.48 & \textbf{1.09} & 0.83 & \textbf{0.52} & 1.20 & 0.35 & 0.49 & 0.54 & 0.84 \\
\text { +Fre-constraints } & 0.85 & 0.86 & \textbf{0.73} & 0.37 & 1.04 & 0.60 & 0.58 & 1.42 & 1.21 & 0.77 & 0.57 & \textbf{1.11} & 0.33 & 0.43 & 0.44 & 0.75 \\
\text { +sampling } & \textbf{0.72} & 0.97 & 0.81 & \textbf{0.36} & 1.01 & \textbf{0.55} & 0.58 & 1.42 & 1.17 & 0.79 & 0.56 & 1.13 & 0.33 & \textbf{0.42} & 0.44 & 0.75 \\
\text { FreNeuS } & 0.81 & \textbf{0.82} & 0.74 & 0.37 & \textbf{0.97} & 0.56 & 0.58 & \textbf{1.40} & 1.19 & \textbf{0.72} & 0.53 & 1.13 & \textbf{0.32} & 0.46 & \textbf{0.42} & \textbf{0.73} \\
\bottomrule
\end{tabular}
\end{adjustbox}
\caption{Ablation study results (chamfer distance) on the DTU (the number in the header indicates the scene ID).}
\label{table:ablation}
\end{table*}

To further validate the effectiveness of our proposed high-frequency guidance method, we compare Fre-NeuS with the global enhancement sampling method, LoD-NeuS\cite{zhuang2023anti}. LoD-NeuS projects four additional rays through the pixel corners, forming a cone-shaped region and sampling its vertices. This approach extends the sampling area for each pixel and uniformly increases the sampling density at each pixel point. We made comparisons on the DTU dataset, as shown in Table \ref{table:Global-compare}, compared to LoD-NeuS, our proposed Fre-NeuS achieves similar results in terms of average error while using less GPU memory with the same training time. This indicates that Fre-NeuS achieves the same effectiveness as global enhanced sampling but with significantly reduced computational cost.

\begin{table}[ht]
\centering
\begin{adjustbox}{scale=0.8}
\small 
\begin{tabular}{l@{\hskip 2pt}cccc}
\toprule
\text{Method} & \text{Train Time} & \text{GPU Memory} & \text{Chamfer distance} \\
\midrule
\text{LoD-NeuS} & $9h$ & $13G$ & $0.72$ \\
\text{FreNeuS (ours)} & $9h$ & $8.7G$ & $0.73$ \\
\bottomrule
\end{tabular}
\end{adjustbox}
\caption{Comparison of GPU Memory, Chamfer distance and Time Consumption with LoD-NeuS, a Global Enhanced Sampling Approach.}
\label{table:Global-compare}
\end{table}

\subsubsection{Generalization validation}
\begin{figure}[t]
  \centering  \includegraphics[width=\linewidth]{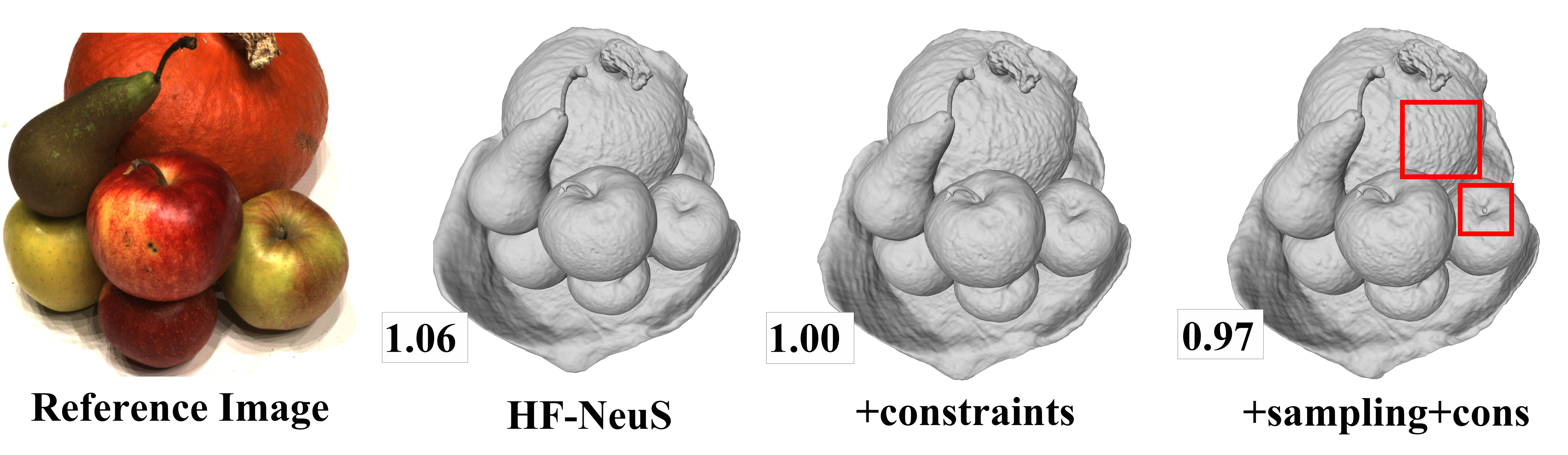}
   \caption{Visual comparisons of improved HF-NeuS after integrating the two proposed modules of FreNeuS. It can be seen that with the addition of the high-frequency constraint mechanism and the high-frequency dynamic sampling strategy, the details of the surface reconstructed by HF-NeuS are clearer. For example, the texture of the pumpkin is more obvious, and the stem of the apple on the far right is also reconstructed.}
   \label{fig:Magra}
\end{figure}

Both modules proposed in FreNeuS: the High-frequency dynamic sampling and the High-frequency constraint mechanisms can be migrated to other NeuS-based models, and they can improve the accuracy of the model in reconstructing local details. To verify this, we conduct experiments using HF-NeuS as an example. Firstly, we add the high-frequency constraints mechanism to HF-NeuS, and ‘+constraints’ in the second row of Table \ref{table:Migration} represents this scheme. 
Then we complete adding two modules to HF-NeuS, the ‘+sampling+cons’ in the third row represents this scheme. From the data in Table \ref{table:Migration}, we can see that both schemes proposed by FreNeuS can improve the accuracy of HF-NeuS. We show the visualisation results in Fig. \ref{fig:Magra}, where we can see that with the addition of the two FreNeuS modules, the local details of the HF-NeuS reconstruction are clearer.
\begin{table}[ht]
\centering
\begin{adjustbox}{max width=\textwidth}
\begin{tabular}{l@{\hskip 2pt}ccccccccccccccccc}
\toprule
\text { Method } & 24 & 37 & 55 & 63 & 65 & 69 & 97 & 105 & 110 & 114 & 118 & 122 & \text { Mean } \\
\midrule
\text { HF-NeuS } & 0.76 & 1.32 & 0.39 & 1.06 & 0.63 & 0.63 & \textbf{1.12} & 0.80 & 1.22 & 0.33	& 0.49 & 0.50 & 0.77 \\
\text {+constraints} & \textbf{0.61} & 1.04 & 0.37 & 1.00 & \textbf{0.59} & \textbf{0.63} & 1.21 & 0.79 & 1.20 & 0.33 & \textbf{0.48} & 0.48 & 0.73 \\
\text {+sampling+cons} & 0.62 & \textbf{0.98} & \textbf{0.36} & \textbf{0.97} & 0.61 & 0.64 & 1.19 & \textbf{0.75} & \textbf{1.20} & \textbf{0.33} & 0.49 & \textbf{0.48} & \textbf{0.71} \\
\bottomrule
\end{tabular}
\end{adjustbox}
\caption{In order to validate the generalizability of the FreNeuS proposed method, a high-frequency constraint mechanism and a high-frequency dynamic sampling strategy are added to the HF-NeuS model. The performance comparison shows that the method proposed by FreNeuS can be adapted to other methods.}
\label{table:Migration}
\end{table}

\subsubsection{Runtime and storage analysis}
To compare the time required for model training, we report the runtime of FreNeuS on the DTU dataset. Neither of the two modules proposed in FreNeuS—the high-frequency constraint mechanism and the high-frequency dynamic sampling strategy—adds extra computational time. Therefore the training time of FreNeuS is close to that of NeuS, both approaching 9 hours on NVIDIA RTX A6000 48GB GPUs. 


In addition, compared to NeuS, FreNeuS adds the high-frequency detection process, which takes about 3 minutes on average on each scene. Therefore, compared to the baseline FreNeuS does not consume too much extra time. Similarly, migrating the FreNeuS method to HF-NeuS does not add too much extra consumption. In terms of scene memory, the average size of all scenes in DTU is used for comparison. As shown in Table \ref{table:time}, the method introduced by FreNeuS also does not add extra memory.
\begin{table}[ht]
\centering
\begin{adjustbox}{scale=0.7}
\begin{tabular}{l@{\hskip 2pt}cccc}
\toprule
\text { Method } & \text { HF-NeuS } & \text { HF+Sampling+constrain } & \text { NeuS } & \text { FreNeuS } \\
\midrule
\text { Time } & $22h$ & $22h$ & $9h$ & $9$ \\
\text {Memory} & $42.15MB$ & $42.5MB$ & $30.13MB$ & $30MB$ \\
\bottomrule
\end{tabular}
\end{adjustbox}
\caption{Memory and time analysis. The approach presented in FreNeuS is not only generalizable but also does not add excessive consumption. The performance of the original model is not degraded, either in terms of training time or memory.}
\label{table:time}
\end{table}


\section{Conclusion}
We present a simple and effective high-frequency guided surface detail reconstruction method, FreNeuS. We observe that local details of the surface can be more clearly represented in the high-frequency image, and therefore utilise the change in pixel gradient to obtain a high-frequency output. FreNeuS then uses the high-frequency information to guide the dynamic sampling of the rays, ensuring that more local detail information is obtained. In addition, to further constrain the reconstruction of texture details, FreNeuS introduces a high-frequency constraint mechanism that constrains the expression of high-frequency information in pixel space. Experiments show that FreNeuS achieves better results in the surface reconstruction task, and it can effectively reproduce surface details. It is worth noting that despite its simplicity, our method has commendable migration properties that make it applicable to any NeuS-based task.












\bibliographystyle{elsarticle-num} 
\bibliography{egbib}






\end{document}